\RequirePackage{fix-cm}
\documentclass[twocolumn]{svjour3}
\pdfoutput=1
\smartqed 
\usepackage{graphicx}
\usepackage{array}
\usepackage{float}
\usepackage[export]{adjustbox}
\usepackage{indentfirst}
\usepackage{amsmath}
\usepackage{hyperref}

\usepackage{latexsym}
\usepackage{times}
\usepackage{epsfig}
\usepackage{epstopdf}
\usepackage{amssymb}
\usepackage{algorithm2e}
\usepackage{gensymb}
\usepackage{xcolor}
\usepackage{cite}
\usepackage{marvosym}
\usepackage{color}
\usepackage{amsmath,amssymb,amsfonts}
\usepackage{algorithmic}
\usepackage{graphicx}
\usepackage{textcomp}
\usepackage{xcolor}
\usepackage[utf8]{inputenc} 
\usepackage[T1]{fontenc}    
\usepackage{hyperref}       
\usepackage{url}            
\usepackage{microtype} 

\begin{document}

\title{ExTTNet: A Deep Learning Algorithm for Extracting Table Texts from Invoice Images}

\author{\href{https://orcid.org/0000-0002-4236-1563}{\includegraphics[scale=0.06]{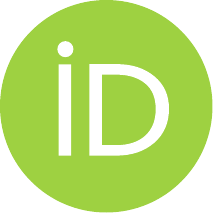}\hspace{1mm}Adem Akdo\u{g}an}\textsuperscript{1}          \and
        \href{https://orcid.org/0000-0002-3236-5595}{\includegraphics[scale=0.06]{orcid.pdf}\hspace{1mm}Murat Kurt}\textsuperscript{2} 
}


\institute{Adem Akdo\u{g}an \at
                \textsuperscript{1} International Computer Institute, Ege University, {I}zmir, Turkey\\
            \email{adem.akdogan92@gmail.com }           
           \and
           Murat Kurt \at
              \textsuperscript{2} International Computer Institute, Ege University, {I}zmir, Turkey\\
              \email{murat.kurt@ege.edu.tr}  
}

\date{Received: date / Accepted: date}

\maketitle

\begin{abstract} In this work, product tables in invoices are obtained autonomously via a deep learning model, which is named as ExTTNet. Firstly, text is obtained from invoice images using Optical Character Recognition (OCR) techniques. Tesseract OCR engine~\cite{smith2007} is used for this process. Afterwards, the number of existing features is increased by using
feature extraction methods to increase the accuracy. Labeling process is done according to whether each text obtained as a result of OCR is a table element or not. In this study, a multilayer artificial neural network model is used. The training has been carried out with an Nvidia RTX 3090 graphics card and taken $162$ minutes. As a result of the training, the F1 score is $0.92$.

\end{abstract}
\newpage
\section{Introduction}\label{intro}

In today's business environment, deep learning technologies are actively utilized in various industry sectors. Crucial roles are played by these technologies in standard company processes like document processing, recording, and digitization. The execution of these tasks can incur not only costs but also time consumption ~\cite{klein2004}. This study introduces the development of a deep learning model designed to rapidly and autonomously extract product tables from invoices, a significant aspect in accounting operations. As a result of this work, a substantial time saving has been achieved in the accounting of invoices.  The study utilized a total of $8794$ invoices from Germany. The open-source optical character recognition engine, Tesseract~\cite{smith2007}, was utilized to extract text information.  This process resulted in obtaining both text information and some corresponding features related to this text. In the subsequent phase, these features are expanded through feature engineering techniques to align with the study's objectives. Following that, a deep learning model is trained using this information to obtain results.

Section~\ref{relwork} delves into the information obtained from literature reviews and similar studies. Section~\ref{methodology} covers processes such as OCR, feature discovery, and model creation. In Section~\ref{ourmodel}, we describe the details of our deep learning model. The obtained results are evaluated in Section~\ref{results}. Finally, conclusions and future work are presented in Section~\ref{futureworks}.

In summary, the novel contributions of this paper are:
\begin{itemize}
    \item  A novel deep learning model (ExTTNet) for extracting table texts from invoice images. 
    \item A utilization of feature engineering techniques to improve accuracy of our ExTTNet.
    \item A detailed validation of our deep learning model (ExTTNet).
\end{itemize}

\section{Previous Studies}\label{relwork}
\subsection{Text-Based Studies}
There are studies related to information extraction from documents that utilize both rule-based and deep learning – machine learning methods. Works such as Intellix ~\cite{schuster2013} and Automatic Indexing ~\cite{esser2012} prominently feature the use of template structures. Therefore, it is more accurate to categorize these studies as rule-based. In the study proposed by Harit and Bansal ~\cite{harit2012}, the information from the header and footer sections of tables is used, resulting in a rule-based method.

While the primary aim of the CloudScan ~\cite{palm2017} study is not to directly obtain tables from invoices, a deep learning algorithm is used to extract other information such as invoice number, invoice date, and invoice amount. Due to similar approaches in data labeling and deep learning modeling, this study is included in the literature review. Similarly, there are machine learning-based studies where different information is obtained from documents, involving processes such as optical character recognition, similar to our work ~\cite{nasiboglu2021}. In our study, unlike the mentioned studies, we aim to predict table elements on invoices. For this purpose, template data on a row basis and relative distance values of texts with respect to each other have been added as features to the dataset.
\subsection{Image-Based Studies}
In addition to text-based approaches, methods that utilize direct image-format documents include techniques for object detection and solutions with rule-based approaches. There are approaches where documents in image format are directly used, as suggested by Gatos et al. ~\cite{gatos2005}. In the method proposed by Gatos et al., horizontal and vertical lines are predicted, followed by obtaining the most suitable section to become a table in a rule-based manner. In the RobusTabNet ~\cite{ma2023} project, a deep learning-based model was developed using direct invoice images. The distinction in our study compared to the aforementioned works lies in using the image only in the initial phase, during the optical character recognition stage, and then conducting training with text information in subsequent stages.

\begin{figure}[t]
	\centering
	\includegraphics[width=0.98\linewidth]{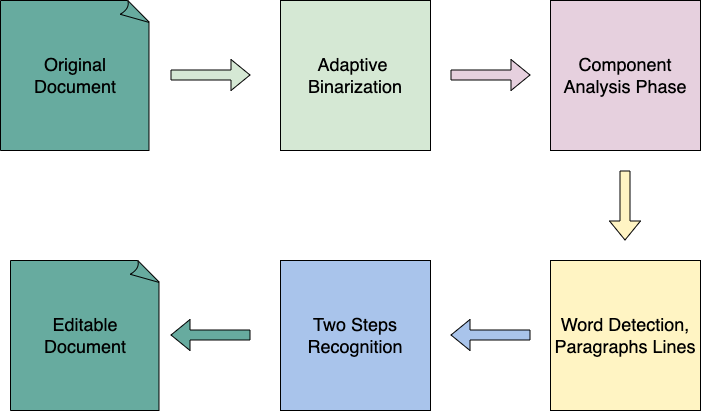}
	\caption{\label{ocr}OCR process.}
\end{figure}

\section{Methodology}\label{methodology}

Firstly, the creation process of documents entering the system is examined. Issues may arise due to errors from the scanner or individuals performing the scanning process when documents are created by passing through the scanner. In such cases, the identified noise on the document is cleaned. It has been observed that some documents are scanned at an angle during the scanning process. As this poses a problem during the OCR stage, documents identified as scanned at an angle are corrected at this stage. Subsequently, the Tesseract OCR engine ~\cite{smith2007} is used to extract text information from the invoice visuals. The general OCR stages are illustrated in Figure \ref{ocr}.

\begin{figure*}[h]
	\centering
	\includegraphics[width=0.98\linewidth]{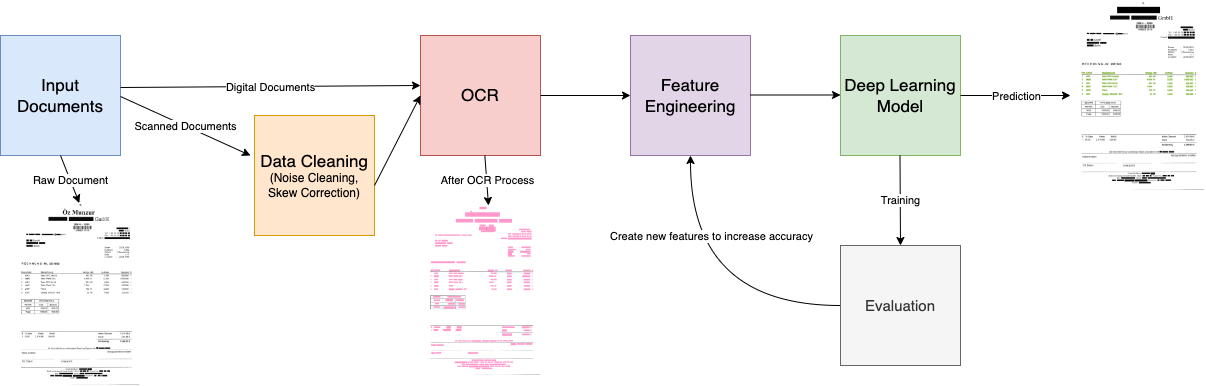}
	\caption{\label{belgehazirlik}Preparation and training procedure for documents employing our deep learning model (ExTTNet).}
\end{figure*}

As a result of this process, each obtained token is labeled as $1$ if it is a table element in the invoice and $0$ if it is not. For each unit obtained through OCR, features such as text height, text width, line number, page width, page height, and the coordinate information of the relevant text are obtained. All detailed features can be examined in Table~\ref{ocrtablosu}. All operations applied to the documents and training processes can be seen in Figure \ref{belgehazirlik}.

\begin{figure*}[t]
	\centering
	\includegraphics[width=0.98\linewidth]{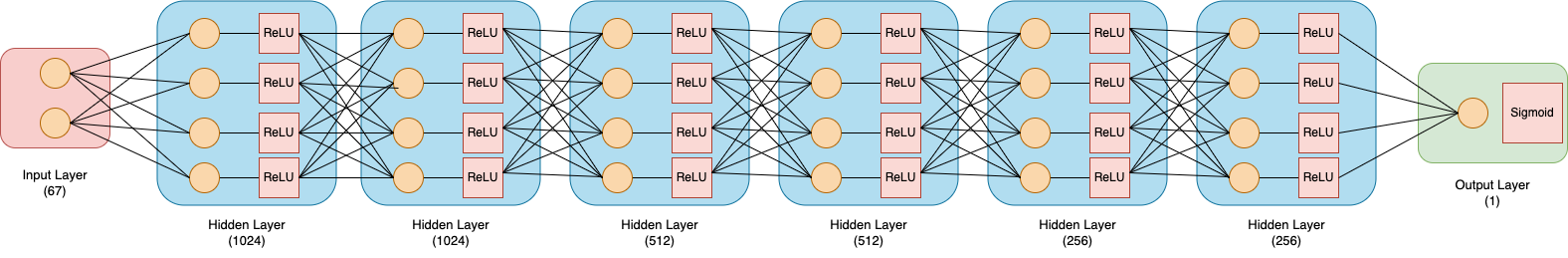}
	\caption{\label{modelmimarisi}Architecture of our deep learning model (ExTTNet).}
\end{figure*}

Additionally, different features are calculated and included in the training data. Some of the features obtained as a result of the processes are specified in Table~\ref{ornektablo}. When calculating these features, the available data is first analyzed in detail. The characteristics of elements outside the tables on the invoice, such as the total block, sub-info, upper info, and address blocks, are examined. Then, the features that best reflect these characteristics are calculated and included in the training data. 

\begin{table}[t]
\centering
\caption{\label{ocrtablosu}Attributes derived from OCR output.}
\begin{tabular}{|p{1.2cm}|p{6.3cm}|}
\hline
\textbf{Attribute Name} & \textbf{Meaning}                               \\ \hline
level                   & Level of the detected unit                    \\ \hline
page\_num               & Page number of the detected unit              \\ \hline
block\_num              & Block number of the detected unit on the page \\ \hline
par\_num                & Paragraph number of the detected unit in the block \\ \hline
line\_num               & Line number of the detected unit in the paragraph \\ \hline
word\_num               & Word number of the detected unit in the line   \\ \hline
left                    & X-coordinate of the top-left corner of the detected unit   \\ \hline
top                     & Y-coordinate of the top-left corner of the detected unit   \\ \hline
width                   & Width of the detected unit                   \\ \hline
height                  & Height of the detected unit                  \\ \hline
conf                    & Confidence percentage of the detected unit   \\ \hline
text                    & Text content of the detected unit             \\ \hline                 
\end{tabular}
\end{table}

\begin{table*}[h]
\caption{\label{ornektablo}Additional attributes utilized during the training process.}
\small 
\begin{tabular}{|p{3cm}|p{10cm}|}
\hline
\textbf{Feature of N-gram} & \textbf{Description}                                                   \\ \hline
RawText                      & Text                                                         \\ \hline
TextPattern                  & Pattern of the text                                                   \\ \hline
BlockNo                      & Information about which block the text is in                           \\ \hline
BlockCharCount               & Total character count within the block                                 \\ \hline
LineWordCount                & Total number of words in the line                                      \\ \hline
BlockWidth                   & Block width                                                           \\ \hline
LineCharCount                & Total character count in the line                                      \\ \hline
IsFirstInt                   & Whether the first character of the text is an Integer                  \\ \hline
BlockWordCount               & Total number of words in the block                                     \\ \hline
PageWidth                    & Page width                                                            \\ \hline
PageHeight                   & Page height                                                           \\ \hline
LeftAlignmentGroup           & Which group the relevant text belongs to with left alignment           \\ \hline
LeftAlignmentCount           & Total number of elements in the left-aligned group where the relevant text is located \\ \hline
RightAlignmentGroup          & Which group the relevant text belongs to with right alignment          \\ \hline
RightAlignmentCount          & Total number of elements in the right-aligned group where the relevant text is located \\ \hline
LineBlockRegex               & Types of all texts on the line where the relevant text is located      \\ \hline
Width                        & Width of the text                                                     \\ \hline
Height                       & Height of the text                                                    \\ \hline
CharCount                    & Total number of letters in the text                                    \\ \hline
Left                         & Left distance of the text (X-coordinate)                               \\ \hline
Top                          & Top distance of the text (Y-coordinate)                                \\ \hline
LeftMargin                   & Proportional left distance of the text to the page                     \\ \hline
TopMargin                    & Proportional top distance of the text to the page                      \\ \hline
FirstQuarter                 & Whether the text is in the first quarter of the page                   \\ \hline
SecondQuarter                & Whether the text is in the second quarter of the page                  \\ \hline
ThirdQuarter                 & Whether the text is in the third quarter of the page                   \\ \hline
FourthQuarter                & Whether the text is in the fourth quarter of the page                  \\ \hline
LineNo                       & Which line the text is in                                             \\ \hline
PageNo                       & Which page the text is on
\\ \hline
\end{tabular}
\end{table*}

When examining the number of aligned texts on the invoice, it is generally observed that the alignment group with the highest number of elements contains table elements. This situation may sometimes be provided by the right alignment group and sometimes by the left alignment group. There are cases where both alignment groups simultaneously provide alignment. In addition, the data analysis conducted revealed that tables are predominantly placed horizontally in the center of the page on invoices. This implies that one of the largest element blocks in the horizontal plane on the page is a table. Another invoice element showing similar characteristics is the footer block. However, this block is generally positioned at the very bottom of the page (in the last quarter of the page) and has a smaller font, hence a smaller text height compared to other elements.

Address blocks are generally located in the first quarter of the page and can be found in the right or left regions of the invoices. The upper info block, containing information such as invoice date, invoice number, delivery date, and customer number, can be situated on the right, left, or center of the page. However, unlike product tables, the types of texts in rows are not constant. All this information is added to the dataset through the attributes shown in Table~\ref{ornektablo}.

To determine the types of texts in a row, each character of the text is examined. If the text consists only of special characters, it is labeled as '?', if it consists only of alphabetic characters, it is labeled as 'W', if it contains only numeric characters, it is labeled as 'N', if it represents a fractional value, it is labeled as 'F', and finally, if the text contains both numeric and alphabetic or special characters together, it is labeled as 'A'. Thus, the rows are patterned based on the types of texts in them. For example, consider a line in an invoice with the text string 'Oktober - Dezember 2019 1,000 ST 70,63 70,63'. In this case, when analyzing the line based on text types, the LineBlockRegex attribute will be 'W ? W N F W F F' ~\cite{potdar2017}.

\section{Our Deep Learning Model (ExTTNet)}\label{ourmodel}
The deep learning model used for training consists of a total of eight layers, including the input layer, output layer, and six different hidden layers, as illustrated in Figure~\ref{modelmimarisi}.

The activation function 'relu' is employed in all layers except the output layer ~\cite{wu2020}. Data labeling is performed based on whether each expression is a table element or not. Therefore, there are two opposite cases. Hence, the 'sigmoid' activation function is used in the final layer.
\section{Results}\label{results}

The learning rate of the deep learning model (ExTTNet) was set to $0.0001$ for this study. Nvidia RTX $3090$ was utilized for training. After preparing the dataset, feature values were normalized. Subsequently, the dataset was divided into training, testing, and validation sets in proportions of $\%70$, $\%20$, and $\%10$, respectively. Following the training process, the success of the deep learning model was evaluated with a previously unused test dataset. F1 score, recall, and precision metrics were employed as success criteria, and the results are presented in Table~\ref{f1tablo} ~\cite{prasanna2018}.

\begin{figure}[t]
	\centering
	\includegraphics[width=0.98\linewidth]{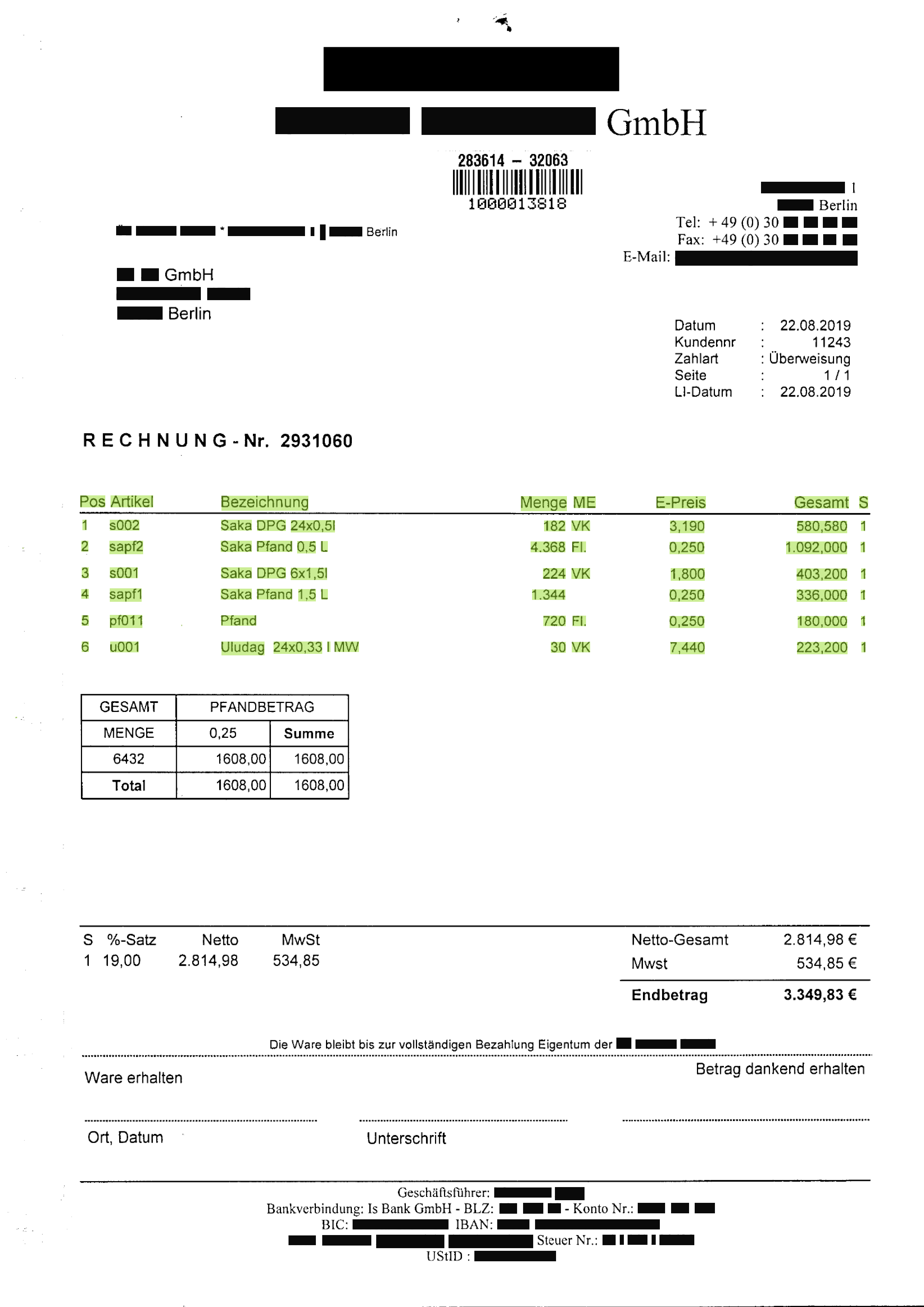}
	\caption{\label{ornekfatura}Sample invoice output generated using our deep learning model (ExTTNet). The text highlighted in green signifies the table elements detected by our deep learning model (ExTTNet).}
\end{figure}

The obtained deep learning model (ExTTNet) predicts whether each text on the invoice is a table element. The prediction results for a sample invoice can be seen in Figure~\ref{ornekfatura}, where highlighted texts (in green) represent table elements.

Predicting each element individually on the invoice has several advantages over predicting the entire table as a block. Firstly, predicting the table as a block from the invoice image requires a high level of accuracy. In the case of a failed prediction, the user would have to manually enter all products on the relevant invoice. However, in the proposed approach of this study, the user only needs to enter the incorrectly predicted table items. Additionally, employing specific algorithmic approaches can further reduce instances of incorrect predictions. Future studies will focus on developing such optimization methods and examining their impact on success.

\begin{table}[t]
\centering
\caption{\label{f1tablo}F1 score, recall, and precision metrics.}
\begin{tabular}{|l|l|l|l|}
\hline
 \textbf{Metrics}                                  & \textbf{Precision} & \textbf{Recall} & \textbf{F1 Score} \\ \hline
\multicolumn{1}{|l|}{0}            & 0.93      & 0.93   & 0.93     \\ \hline
\multicolumn{1}{|l|}{1}            & 0.92      & 0.91   & 0.91     \\ \hline
\multicolumn{1}{|l|}{accuracy}     &           &        & 0.92     \\ \hline
\multicolumn{1}{|l|}{macro avg}    & 0.93      & 0.92   & 0.92     \\ \hline
\multicolumn{1}{|l|}{weighted avg} & 0.93      & 0.93   & 0.93     \\ \hline
\end{tabular}
\end{table}

\section{Conclusions and Future Work}\label{futureworks}

In this study, we achieved the autonomous extraction of product tables from invoices in image format using a deep learning model, which is called as ExTTNet. Particularly in companies with intensive accounting operations, this artificial intelligence model (ExTTNet) is expected to assist users in completing accounting processes more efficiently.  

Based on the findings of ExTTNet, there are plans to make enhancements in future research with the aim of improving the effectiveness of the deep learning model. The primary improvement under consideration is the expanded utilization of images, not just in the text extraction phase but also within the deep learning model itself. In document image formats like invoices, certain details such as shapes, symbols, and barcode configurations, which are not obtainable through OCR, are present. Currently, this information cannot be leveraged in the training phase as it doesn't reach the text generation component. However, if images can be incorporated into the training phase, this overlooked information will be taken into account. Additionally, there is potential for upgrading and optimizing ExTTNet by incorporating state-of-the-art methods and
techniques~\cite{Gok2023SIU,Azadvatan2024arXiv}.  

We've carefully crafted a set of different features to help predict and understand table elements These features play a key role in anticipating and comprehending the elements found in tables. To make our predictions even more accurate and reliable, we're looking into using advanced techniques to fine-tune and improve these features. This strategic improvement will ensure a better grasp of the complex relationships and patterns present in the elements of the table.

Additionally, at the initial stage, image cleaning and correction processes are applied, and ongoing improvements in these procedures are in the planning stage. These enhancements will bolster Tesseract's performance, leading to a more precise extraction of text in the OCR process. Consequently, the feature engineering segment will receive higher-quality data, further contributing to overall success.

Furthermore, we are also interested in implementing our novel deep learning algorithm for representing Bidirectional Reflectance Distribution Functions (BRDFs)~\cite{Ozturk2006EGUK,Kurt2007MScThesis,Ozturk2008CG,Kurt2008SIGGRAPHCG,Kurt2009SIGGRAPHCG,Kurt2010SIGGRAPHCG,Ozturk2010GraphiCon,Szecsi2010SCCG,Ozturk2010CGF,Bigili2011CGF,Bilgili2012SCCG,Ergun2012SCCG,Toral2014SIU,Tongbuasirilai2017ICCVW,Kurt2019DEU,Tongbuasirilai2020TVC,Akleman2024arXiv}, Bidirectional Scattering Distribution Functions (BSDFs)~\cite{WKB12,Ward2014MAM,Kurt2014WLRS,Kurt2016SIGGRAPH,Kurt2017MAM,Kurt2018DEU}, Bidirectional Surface Scattering Reflectance Distribution Functions (BSSRDFs)~\cite{Kurt2013TPCG,Kurt2013EGSR,Kurt2014PhDThesis,Onel2019PL,Kurt2020MAM,Kurt2021TVC,Yildirim2024arXiv} and multi-layered materials~\cite{WKB12,Kurt2016SIGGRAPH,Mir2022DEU} in computer graphics.


%
%



\bibliographystyle{spmpsci}      
\bibliography{arXiv24_ExTTNET_References}   

\end{document}